\documentclass{article}

\usepackage[numbers]{natbib}  % Load early with options

\usepackage[final, nonatbib, sglblindworkshop]{neurips_2025}
\usepackage{amsmath}
\usepackage{graphicx}
\usepackage{microtype}
\usepackage{graphicx}
\usepackage{subfigure}
\usepackage{booktabs} % for professional tables
\usepackage{wrapfig}

\usepackage{hyperref}

\usepackage{caption}
\usepackage{placeins}
\usepackage{amsmath}
\usepackage{amssymb}
\usepackage{mathtools}
\usepackage{amsthm}

\usepackage[capitalize,noabbrev]{cleveref}

\usepackage{xcolor}         % colors
\usepackage{enumitem}
\usepackage{float}
\usepackage{arydshln}
\usepackage{multirow}

\newcommand{\ie}{\textit{i}.\textit{e}., }
\newcommand{\eg}{\textit{e}.\textit{g}., }

\DeclareMathOperator*{\softmax}{\operatorname{softmax}}

\DeclareMathOperator*{\att}{\operatorname{Attention}}
\DeclareMathOperator*{\multihead}{\operatorname{MultiHead}}

\newcommand{\method}{Solar-GECO}

\title{\method: Perovskite Solar Cell Property Prediction with Geometric-Aware Co-Attention}

\author{%
  Lucas Li$^1$,~ Jean-Baptiste Puel$^{2,3}$,~ Florence Carton$^1$,~ Dounya Barrit$^{1,3}$,~ Jhony H. Giraldo$^4$ \\ \\
  $^1$TotalEnergies OneTech, 91120 Palaiseau, France\\ 
  $^2$EDF, Electricité de France, 91120 Palaiseau, France \\ 
  $^3$IPVF, Institut Photovoltaïque d’Ile-de-France, 91120 Palaiseau, France \\
  $^4$LTCI, Télécom Paris, Institut Polytechnique de Paris, 91120 Palaiseau, France \\
}

\workshoptitle{AI4Mat}

\begin{document}

\maketitle
\begin{abstract}
Perovskite solar cells are promising candidates for next-generation photovoltaics.
However, their performance as multi-scale devices is determined by complex interactions between their constituent layers.
This creates a vast combinatorial space of possible materials and device architectures, making the conventional experimental-based screening process slow and expensive.
Machine learning models try to address this problem, but they only focus on individual material properties or neglect the important geometric information of the perovskite crystal.
To address this problem, we propose to predict perovskite \textbf{solar} cell power conversion efficiency with a \textbf{ge}ometric-aware \textbf{co}-attention (\method) model.
\method~combines a geometric graph neural network (GNN)---that directly encodes the atomic structure of the perovskite absorber---with language model embeddings that process the textual strings representing the chemical compounds of the transport layers and other device components.
\method~also integrates a co-attention module to capture intra-layer dependencies and inter-layer interactions, while a probabilistic regression head predicts both power conversion efficiency (PCE) and its associated uncertainty.
\method~achieves state-of-the-art performance, significantly outperforming several baselines, reducing the mean absolute error (MAE) for PCE prediction from $3.066$ to $2.936$ compared to semantic GNN (the previous state-of-the-art model). 
\method~demonstrates that integrating geometric and textual information provides a more powerful and accurate framework for PCE prediction.
%for accelerating the design of complex, multi-component materials systems.
\end{abstract}

\section{Introduction}

Machine learning has transformed materials science by enabling fast prediction of isolated material properties, such as bandgap, formation energy, or carrier mobility \cite{Dunn2020,Choudhary2024,Lee2023}.
While these efforts have accelerated the discovery of promising compounds, they often focus on single-scale problems where the target property is intrinsic to the material itself.
However, in real-world applications, optimal device performance arises from the coupled behavior of multiple components across diverse scales \cite{Moosavi2020}.
For complex optoelectronic devices such as perovskite solar cells, efficiency depends not only on the properties of the perovskite absorber, but also on the underlying interactions between transport layers, electrodes, and their interfaces \cite{Stolterfoht2019}. 
This multiscale interdependence poses challenges that go beyond conventional single-material property prediction \cite{Aneesh2025}.

Perovskite solar cells have achieved rapid progress in laboratory efficiency \cite{Guo2023}, but their commercialization encounters a fundamental bottleneck: the extensive combinatorial space of potential materials, device architectures, and processing conditions.
Each device layer---such as the hole transport layer (HTL), electron transport layer (ETL), and encapsulation---can be produced using several prospective materials, each with its own variants in stoichiometry, morphology, and processing \cite{liu2018hole,liu2022machine}.
Therefore, the total number of possible configurations grows exponentially.
Physics-guided and intuition-driven design cannot explore this design space at the pace required by current innovation cycles \cite{Unold2022, Dale2023}.
The result is a gap between the growing diversity of prospect materials and the rate at which optimal full-device architectures can be identified.

Conventional development pipelines often adopt a sequential strategy: first identifying high-performing materials in isolation, then attempting to integrate them into full devices \cite{mao2025comprehensive}.
However, this approach can be misleading because many promising combinations only exhibit their full potential when considered holistically, as interactions between layers can enhance---or severely degrade---performance \cite{liang2022accelerating}.
This phenomenon is not unique to perovskite solar cells; similar issues arise in other multicomponent systems such as batteries, catalysts, and thermoelectrics \cite{Charalambous2024}.
%, where data availability and quality vary widely across scales.
Bridging the gap between layer-level and full-device performance prediction requires models that can represent both intra-layer properties and inter-layer relationships within the same device.

\begin{wrapfigure}{r}{0.55\textwidth}
    %\vspace{-5pt}
    \centering
    \includegraphics[width=0.55\textwidth]{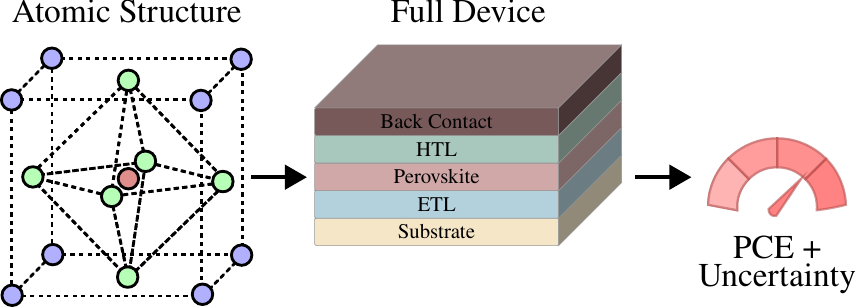}
    \caption{Our model integrates the atomic structure of the perovskite absorber (left) with the device context of stacked layers (center) through a co-attention module. The fused representation is used to predict the power conversion efficiency (PCE) of the device and its associated uncertainty (right).}
    \label{fig:teaser}
    \vspace{-10pt}
\end{wrapfigure}

In this work, we propose to predict perovskite \textbf{solar} cell power conversion efficiency with a \textbf{ge}ometric-aware \textbf{co}-attention (\method) model.
\method~is a hybrid algorithm where we explicitly integrate crystal-level information from the perovskite layer with device-level architectural context.
Unlike prior work on semantic device graphs \cite{Aneesh2025}, which rely solely on text embeddings from a large language model (LLM), our method processes the crystal structure of the perovskite absorber directly with a geometric graph neural network (GNN).
%uses large language models (LLMs) to embed all layers in the device and employs graph neural networks (GNNs) to represent the device as a whole, our method processes the crystal structure of the perovskite absorber directly with a geometric GNN.
This allows the extraction of physically grounded features from the crystal structure as shown in Figure \ref{fig:teaser}.
Other layers in the device are encoded using LLM-derived molecular embeddings.

%We propose \method, a geometric-aware co-attention model that integrates crystal-level information from the perovskite absorber with device-level context. Unlike semantic device graph approaches \cite{Aneesh2025}, which rely solely on text embeddings, our method directly encodes the absorber’s crystal structure with a geometric GNN while representing the other layers with LLM-derived embeddings.

In \method, we also introduce self-attention and cross-attention mechanisms to jointly model intra-layer dependencies and inter-layer interactions, capturing how the interaction between atoms in the perovskite layer and the context layer of the device propagates to the device performance.
Significant variability in the PCE may derive from fabrication process sensitivities, including parameters like humidity and annealing temperature, and unmodeled factors such as human error, material insolubility, and poor wettability.
To account for some of this inherent uncertainty, our model predicts the PCE using a Gaussian negative log-likelihood (NLL) loss function.
We evaluate \method~on a curated subset of the Perovskite Database \cite{Jacobsson2022} and the Materials Project \cite{Jain2013}, achieving state-of-the-art performance.
Our main contributions are as follows:
\begin{itemize}[leftmargin=0.5cm]
    \item We propose a novel model, \method, that combines crystal-level geometric GNN encoding of the perovskite absorber with LLM-based molecular embeddings for the context device layers.
    \item We introduce a co-attention module that combines self-attention within layers and cross-attention across layers, enabling mutual refinement of graph and text representations.
    %We introduce a co-attention module that integrates self-attention for intra-layer dependencies and cross-attention for inter-layer interactions, allowing mutual refinement of graph and text representations.
    \item \method~models uncertainty in PCE predictions by training with a Gaussian NLL loss.
    \item Our model achieves state-of-the-art for PCE prediction in perovskite solar devices.
\end{itemize}

%and the Materials Project \cite{Jain2013}, achieving state-of-the-art predictive accuracy and demonstrating the benefits of integrating crystal-level GNN encodings into holistic device modeling.

\section{Related Work}

\textbf{Property prediction in materials.}
Several machine learning approaches have been applied to predict material properties.
%for perovskite solar cells.
Early efforts focused on composition-to-property regression using algorithms such as gradient boosting regression tree, kernel ridge regression, and support vector machines to estimate bandgaps and stability metrics \cite{Parikh2022,Im2019,Li2019}.
More recent methodologies include structure-based models that only rely on chemical composition.
For example, CrabNet \cite{Wang2021crabnet} uses a Transformer-based architecture to learn the relative importance of elements for target property prediction, without requiring any explicit structure information.
With the rise of geometric deep learning, GNNs have become the dominant paradigm for predicting material properties from structure, as they can easily represent atomic interactions and capture complex composition–structure–property relationships \cite{Gilmer2017, Schutt2018}.
In particular, geometric GNNs, which incorporate 3D atomic coordinates and respect physical symmetries such as E$(3)$-invariance/equivariance, have shown superior data efficiency and generalization across materials domains \cite{duval2023hitchhiker}.
For crystalline materials, specialized architectures such as Matformer \cite{Yan2022} explicitly encode periodic boundary conditions and lattice geometry, enabling improved prediction of electronic and structural properties.

% In parallel to structure-based methods, models that only rely on chemical composition have also demonstrated significant success.
% For example, CrabNet \cite{Wang2021crabnet} uses a Transformer based architecture to learn the relative importance of elements for target property prediction, without requiring any explicit structure information.

%CrabNet \cite{Wang2021crabnet}, which uses a Transformer based architecture to learn the relative importance of elements for target property prediction, without requiring any explicit structure information.
{\color{black}
\textbf{Multimodal and 
attention based fusion.}
Traditional early- and late-fusion methods rely on explicit feature concatenation or decision-level averaging, which limits their ability to capture rich cross-modal interactions \citep{ngiam2011multimodal, poria2017context}. 
More recently, attention-based multimodal architectures have been proposed to model interactions directly across modalities\citep{tsai2019multimodal}. 
}

\textbf{Device-level perovskite property prediction.}
While most prior work targets isolated material properties, the recent \emph{semantic device graphs} method \cite{Aneesh2025} models the entire device stack, representing each layer and its interfaces as a heterogeneous graph.
LLMs are used to embed the textual strings representing the chemical compounds in the device, and a GNN is applied to capture inter-layer relationships.
This holistic view enables the identification of high-performing layer combinations that might be overlooked by sequential screening of individual materials.
However, this method represents the perovskite absorber only through text embeddings, ignoring its crystallographic structure.

%treats the perovskite absorber layer only via text-based embeddings, without leveraging its full crystallographic structure.

Contrary to previous works, we explicitly include the atomic structure information in the prediction of PCE in perovskite devices.
To this end, we directly process the perovskite’s atomic structure with a geometric GNN, incorporating physically grounded features that improve PCE prediction accuracy.

% Our work builds on these developments by combining crystal-level geometric GNN encoding for the perovskite absorber with LLM-based embeddings for the transport layers, enabling a unified intra- and inter-layer representation.
% By directly processing the absorber’s atomic structure, our approach complements prior semantic device graph models and incorporates physically grounded features that improve predictive accuracy for power conversion efficiency on the Perovskite Database \cite{Jacobsson2021}.

\section{Methodology: \method}

\subsection{Preliminaries}

\textbf{Notations}.
In this work, we denote sets using calligraphic letters (\eg $\mathcal{V}$), with their cardinality written as $|\mathcal{V}|$.
Bold uppercase letters (\eg $\mathbf{H}$) represent matrices, and bold lowercase letters (\eg $\mathbf{x}$) represent vectors.
The transpose operator is denoted by $(\cdot)^{\top}$ and concatenation by $[~\cdot~;~\cdot~]$.

% \textbf{Basic definitions}.
% We represent a graph as \( G = (\mathcal{V}, \mathcal{E}) \), where \( \mathcal{V} \) is the set of vertices and \( \mathcal{E} \subseteq \mathcal{V} \times \mathcal{V} \) is the set of edges. 
%Node features are stored in a matrix \( \mathbf{X}_{\mathcal{V}} \in \mathbb{R}^{|\mathcal{V}| \times m} \) and \( \mathbf{F}_{\mathcal{E}} \in \mathbb{R}^{|\mathcal{E}| \times l} \), where \( m \) and \( l \) denote the respective feature dimensions.
%Each edge \( e \in \mathcal{E} \) is an ordered pair \( (u, v) \) representing a connection between vertices \( u \) and \( v \).

\textbf{Perovskite solar device}.
A perovskite solar cell is a multi-layered photovoltaic device.
%its overall performance is not just determined by the quality of the central perovskite material, but by the complex interplay between all constituent layers.
A typical architecture consists of five different layers as shown in the center of Figure \ref{fig:teaser}.
The substrate is the foundational base, typically glass with an FTO (fluorine-doped tin oxide) or ITO (indium tin oxide) layer that acts as the front electrode.
We also have two charge transport layers: \textit{i}) the ETL which extracts electrons, and \textit{ii}) the HTL that extracts the holes.
In the core of the device, we have the perovskite absorber, which absorbs sunlight to generate electron-hole pairs.
Finally, the back contact is an electrode that completes the electrical circuit.

% Substrate: The foundational base, which acts as the front electrode.
% Charge Transport Layers: The electron transport layer (ETL) extracts electrons, and the hole transport layer (HTL) extracts holes.
% Perovskite Absorber: It absorbs sunlight to generate electron-hole pairs.
% Back Contact: A final electrode,that completes the electrical circuit.

\textbf{Graph representation of crystals}.
In materials science, a crystal structure can be naturally represented as a graph $G = (\mathcal{V}, \mathcal{E})$, where the set of vertices $\mathcal{V}$ corresponds to the atoms in the crystal unit cell, and the set of edges $\mathcal{E}$ represents the connections or bonds between them.
Typically, an edge exists between two atoms if their interatomic distance is within a specified cutoff radius.
The features of each node can include atomic properties (\eg electronegativity, atomic mass), and edge features can represent the distances between connected atoms.

\textbf{Geometric graph neural networks}.
GNNs are deep learning models designed to operate on graph-structured data by learning node representations through iterative message passing \cite{scarselli2009graph}.
For 3D atomic systems---such as molecules, proteins, and crystalline materials---a specialized class of models known as geometric GNNs has become the state-of-the-art architecture \cite{duval2023hitchhiker}.

These models represent atomic systems as geometric graphs, where atoms are nodes endowed with coordinates in 3D Euclidean space ($\mathbb{R}^3$). 
The core principle of a geometric GNN is the incorporation of fundamental physical symmetries as a powerful inductive bias. 
Specifically, the learned representations are constrained to respect the symmetries of the Euclidean group E$(3)$, which encompasses all rotations, reflections, and translations, as well as permutation invariance with respect to node indexing.
This property is known as E$(3)$-equivariance.
Formally, if the input atomic coordinates are transformed by an operation $g(\cdot) \in \text{E}(3)$, the model's feature vectors at each layer must also transform predictably according to a corresponding group representation of $g(\cdot)$.
% A special case of this E$(3)$-invariance is the total potential energy of a system, which remains unchanged under these transformations.
\textcolor{black}{A special case of equivariance is invariance, where the output remains unchanged under these transformations , such as the total potential energy of a system, which is invariant to rotations and translations.}

%By embedding these physical constraints directly into the model architecture, geometric GNNs achieve superior data efficiency and generalization for a wide range of scientific applications, from molecular simulation and drug discovery to materials design.

%\textcolor{red}{TODO Jhony: Check geometric GNN part after correction.}

\textbf{Attention mechanism}.
\method~uses the multi-head scaled dot-product attention mechanism \cite{vaswani2017attention} to process intra- and inter-layer information within the perovskite device.
%Each fusion layer is built upon the Multi-Head Scaled Dot-Product Attention mechanism.
In general terms, this mechanism weighs the importance of different elements in a sequence or set of tokens.
An attention function is a mapping of a query and a set of key-value pairs to an output.
The output is computed as a weighted sum of the values, where the weight assigned to each value is computed by a compatibility function of the query with the corresponding key.

Formally, given a query matrix $\mathbf{Q} \in \mathbb{R}^{N\times d_k}$, a key matrix $\mathbf{K}\in \mathbb{R}^{N\times d_k}$, and a value matrix $\mathbf{V}\in \mathbb{R}^{N\times d_v}$, the scaled dot-product attention is calculated as:
\begin{equation}
\att(\mathbf{Q}, \mathbf{K}, \mathbf{V}) = \softmax\left(\frac{\mathbf{Q}\mathbf{K}^\top}{\sqrt{d_k}}\right)\mathbf{V},
\end{equation}
where $d_k$ is the dimension of the queries and keys, $d_v$ is the dimension of the values, and $N$ is the number of tokens.
The matrix product $\mathbf{Q}\mathbf{K}^\top$ computes the dot product between every possible pair of rows in $\mathbf{Q}$ and $\mathbf{K}$.
In other words, this inner product computes the similarity between each query and all keys.
%The dot product $\mathbf{Q}\mathbf{K}^\top$ computes the similarity between each query and all keys. 
The result is scaled by $1/\sqrt{d_k}$, and a $\softmax$ function is applied to obtain the so-called attention weights.
Finally, we weigh the values matrix with the attention weights, which assign different levels of importance (attention) to the set of values.
Our model uses multi-head attention, which applies this mechanism multiple times in parallel with different, learned linear projections of $\mathbf{Q}$, $\mathbf{K}$, and $\mathbf{V}$, defined as follows:
\begin{equation}
\begin{split}
    \multihead(\mathbf{Q}, \mathbf{K}, \mathbf{V}) &= \left[\mathbf{M}_1; \dots; \mathbf{M}_h\right]\mathbf{W}^O, \\
    \text{where ~} \mathbf{M}_i &= \att(\mathbf{Q}\mathbf{W}_i^Q, \mathbf{K}\mathbf{W}_i^K, \mathbf{V}\mathbf{W}_i^V).
\end{split}
% \begin{split}
%     \multihead(\mathbf{Q}, \mathbf{K}, \mathbf{V}) &= \concat(\text{head}_1, \dots, \text{head}_h)\mathbf{W}^O \\
%     \text{where} \quad \text{head}_i &= \att(\mathbf{Q}\mathbf{W}_i^Q, \mathbf{K}\mathbf{W}_i^K, \mathbf{V}\mathbf{W}_i^V).
% \end{split}
\end{equation}
Here, $h$ is the number of heads, and $\mathbf{W}_i^Q\in\mathbb{R}^{d_{\text{model}}\times d_k}, \mathbf{W}_i^K\in\mathbb{R}^{d_{\text{model}}\times d_k}, \mathbf{W}_i^V\in\mathbb{R}^{d_{\text{model}}\times d_v}, \mathbf{W}^O\in\mathbb{R}^{h d_v \times d_{\text{model}}}$ are learnable projection matrices for the head $i$, where $d_{\text{model}}$ is the dimensionality of the input embeddings.
The multi-head attention allows the model to jointly attend to information from different representation subspaces at different positions.

\begin{figure}
    \centering
    \includegraphics[width=\textwidth]{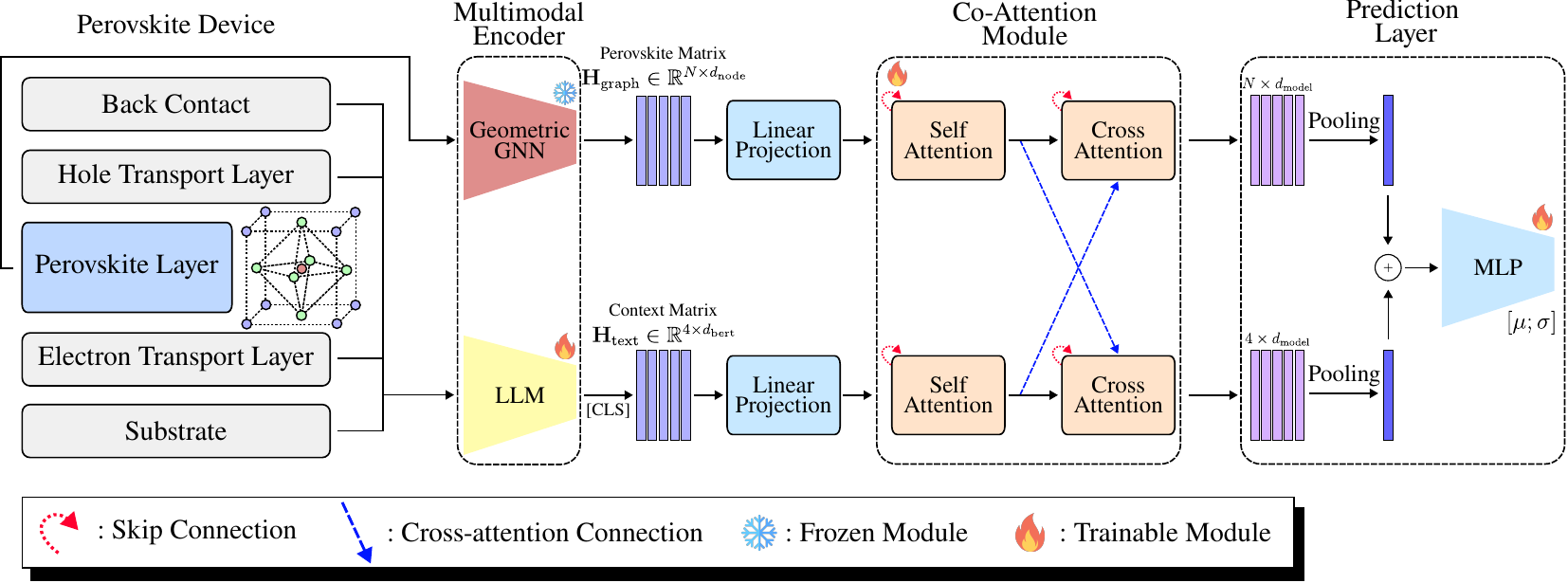}
    \caption{Overview of \method. Our model combines a geometric GNN encoder for the perovskite absorber’s 3D atomic structure with a device text encoder (LLM) for chemical descriptions of the substrate, ETL, HTL, and back contact. Their outputs are fused in a co-attention module with self- and cross-attention layers to model intra-layer dependencies and inter-layer interactions. Pooled features are concatenated and passed to a probabilistic regression head that predicts the mean and variance of PCE, trained with a Gaussian negative log-likelihood loss.}
    \label{fig:pipeline} 
\end{figure}

\textbf{Problem definition}.
Given a dataset of device-PCE pairs $\mathcal{D}=\{ \mathbf{x}_i, \mathbf{y}_i \}_{i=1}^M$, where $\mathbf{x}_i$ is the $i$th perovskite solar device and $\mathbf{y}_i$ is its PCE, our objective is to train a model $F_{\boldsymbol{\theta}}$ with parameters $\boldsymbol{\theta}$ to predict the PCE of each device.
In practice, since our datasets come from experimental work, one single $\mathbf{x}_i$ might have different values of PCE.
In this work, we use neural networks to parametrize $F_{\boldsymbol{\theta}}$.

%, \ie there exists some $\mathbf{x}_i~|~\mathbf{y}_i = \{ \mathbf{y}^{(1)}_i,\dots,\mathbf{y}^{(P)}_i\}$

\subsection{Overview}

Figure \ref{fig:pipeline} shows the pipeline of our method.
\method~is designed to predict the PCE of perovskite cells by integrating information from the perovskite's crystal structure and the device's text representations.
The pipeline consists of three main stages.
First, we employ two specialized, pre-trained encoders to independently extract features: a crystal graph convolutional neural network (CGCNN) \cite{Xie2018} processes the atomic structure of the perovskite, while a MaterialsBERT model \cite{Gupta2022} processes the text string representing the chemical compounds of the device's functional layers. 
Second, these distinct feature sets are fed into a multi-layer co-attention module, which processes the interdependent relationships between the material's structural properties and the device's architectural context.
Finally, the fused, context-aware representations are passed to a probabilistic regression head that predicts the PCE and the inherent uncertainty of its prediction, optimized using a Gaussian NLL loss.

%\subsection{Preliminaries}

\subsection{Multi-Modal Feature Extraction}

The initial step of our model involves extracting high-quality representations from the two distinct input modalities: the perovskite's crystal structure and the device's text representations.

\textbf{Crystal graph encoder}.
We use a pre-trained CGCNN \cite{Xie2018} to encode the perovskite's atomic structure.
Given a crystal graph $G$, the input representation for each node is obtained based on its atomic number.
CGCNN applies a series of graph convolution operations where each atom's feature vector is updated based on the features of its neighboring atoms and the distances between them.
This process yields a feature vector for each atom (node) in the graph, resulting in a sequence of node features $\mathbf{H}_{\text{graph}} \in \mathbb{R}^{N \times d_{\text{node}}}$, where $N$ is the number of atoms in the perovskite layer.

\textbf{Device text encoder}.
For the device architecture, we use a pre-trained MaterialsBERT model \cite{Gupta2022} to encode the textual strings representing the chemical compounds of the four context layers: the substrate, ETL, HTL, and back contact.
The input string of each layer is individually tokenized and processed by the model.
Following standard practice \cite{Devlin2019}, we extract the hidden state of the special [CLS] token from the encoder as the representative embedding for each component.
These four embedding vectors are then stacked to form a sequence of text-based features $\mathbf{H}_{\text{text}} \in \mathbb{R}^{4 \times d_{\text{bert}}}$.

\subsection{Co-Attention Fusion Module}

To effectively model the relationships between the perovskite and the context layers of the device, we introduce a co-attention fusion module.
To this end, we define a co-attention fusion layer by stacking a self-attention and a cross-attention operation for each branch (graph and text) as shown in Figure \ref{fig:pipeline}.
This module enables the graph and text representations to be mutually refined.

%To model relationships between the perovskite and device layers, we introduce a co-attention module that stacks self- and cross-attention for graph and text branches, enabling mutual refinement of their representations.

\textbf{Intra-modal self-attention}.
The first step within each fusion layer is to refine the representation of each modality independently using self-attention.
In this case, the queries, keys, and values are all derived from the same input sequence, allowing each element (atom or text token) to attend to all other elements within its modality.
For the graph representation, let $\mathbf{H}^{(l-1)}_{\text{graph}}$ be the output of the previous $l-1$ co-attention fusion layer of the graph branch, where $1 \leq l \leq L$, and $\mathbf{H}^{(0)}_{\text{graph}}=\mathbf{H}_{\text{graph}}\mathbf{W}_\text{graph}$, \ie the output of the CGCNN after linear projection with $\mathbf{W}_{\text{graph}} \in \mathbb{R}^{d_{\text{node}} \times d_{\text{model}}}$.
The self-attention block is then defined as:
% the input to the graph self-attention
% denoted as $\mathbf{H}^{(l-1)}_{\text{graph}}$ , where $l$ is the Co-attention layers (from 1 to $L$), $\mathbf{H}^{(0)}_{\text{graph}}$ is the sequence of node features produced by the crystal graph encoder after the initial linear projection :
\begin{equation}
    \mathbf{H'}^{(l)}_{\text{graph}} = \gamma\left(\mathbf{H}^{(l-1)}_{\text{graph}} + \multihead \left(\mathbf{H}^{(l-1)}_{\text{graph}},\mathbf{H}^{(l-1)}_{\text{graph}},\mathbf{H}^{(l-1)}_{\text{graph}}\right)\right),
\end{equation}
where $\gamma(\cdot)$ is a normalization function.
An analogous operation is performed for the text branch, where $\mathbf{H}^{(0)}_{\text{text}}=\mathbf{H}_{\text{text}}\mathbf{W}_\text{text}$, \ie the output of the BERT model after linear projection.
This self-attention step updates the embedding for each element by aggregating information from all other elements within the same modality before the cross-attention module.

\textbf{Inter-modal cross-attention}.
The second step is the core of the fusion process: inter-modal bidirectional cross-attention.
Here, representations from the two modalities query each other.
To update the graph representation, its self-attended features $\mathbf{H}'_{\text{graph}}$ act as the queries, while the self-attended text features $\mathbf{H}'_{\text{text}}$ provide the keys and values.
This allows each atom's representation to be updated with information from the entire device stack.
The operation is thus defined as:
\begin{equation}
    \mathbf{H}^{(l)}_{\text{graph}} = \gamma\left(\mathbf{H'}^{(l)}_{\text{graph}} + \multihead\left(\mathbf{H'}^{(l)}_{\text{graph}}, \mathbf{H'}^{(l)}_{\text{text}}, \mathbf{H'}^{(l)}_{\text{text}}\right)\right).
\end{equation}
Similarly, for the text branch, we have:
\begin{equation}
    \mathbf{H}^{(l)}_{\text{text}} = \gamma\left(\mathbf{H'}^{(l)}_{\text{text}} + \multihead\left(\mathbf{H'}^{(l)}_{\text{text}}, \mathbf{H'}^{(l)}_{\text{graph}}, \mathbf{H'}^{(l)}_{\text{graph}}\right)\right).
\end{equation}
The entire process of self-attention followed by cross-attention is repeated for $L$ layers, allowing for progressively deeper integration of the multi-modal information.
This architecture allows the model to iteratively learn which atoms in the crystal are most relevant to specific device layers, and conversely, which device layers are most influenced by specific crystal structure. 

% \begin{equation}
%     \mathbf{H}^{(l)}_{\text{graph}} = \gamma\left(\mathbf{H'}^{(l)}_{\text{graph}} + \text{CrossAttention}\left(\text{Q}=\mathbf{H'}^{(l)}_{\text{graph}}, \text{KV}=\mathbf{H'}^{(l)}_{\text{text}}\right)\right)
% \end{equation}

% \begin{equation}
%     \mathbf{H}^{(l)}_{\text{text}} = \text{LayerNorm}\left(\mathbf{H'}^{(l)}_{\text{text}} + \text{CrossAttention}\left(\text{Q}=\mathbf{H'}^{(l)}_{\text{text}}, \text{KV}=\mathbf{H'}^{(l)}_{\text{graph}}\right)\right)
% \end{equation}    

% \textcolor{red}{Should we say that we add skip connections for both, the self and cross-attention layers?}

% \textcolor{blue}{Indeed,I modified equations before}

%\textcolor{blue}{TODO Lucas: Prediction section}

\subsection{Prediction Layer}

%The fabrication process of perovskite solar cells is a complex process, devices with identical material compositions and architectures often have a distribution of power conversion efficiency rather than a single.
% To account for this inherent uncertainty in PCE prediction, \method~is designed to predict a probability distribution that represents the likely range of PCEs.

\textbf{Layer-wise fusion and pooling}.
After the final cross-attention layer, we apply masked average pooling on the graph features $\mathbf{H}^{(L)}_{\text{graph}} \in \mathbb{R}^{N \times d_{\text{model}}}$ and text features $\mathbf{H}^{(L)}_{\text{text}} \in \mathbb{R}^{4 \times d_{\text{model}}}$.
Therefore, we obtain single fixed-size vectors $\mathbf{v}_{\text{graph}} \in \mathbb{R}^{d_{\text{model}}}$ and $\mathbf{v}_{\text{text}} \in \mathbb{R}^{d_{\text{model}}}$ for the  perovskite and context layers, respectively.
These features are concatenated to form the final feature vector, \ie $\mathbf{v}_{\text{final}} = [\mathbf{v}_{\text{graph}} ; \mathbf{v}_{\text{text}}] \in \mathbb{R}^{2d_{\text{model}}}$, which is then passed to the regression head.

% to sequences of graph and node features, $\mathbf{H}_{\text{graph}} \in \mathbb{R}^{N \times d_{\text{model}}}$, and text token features, $\mathbf{H}_{\text{text}} \in \mathbb{R}^{M \times d_{\text{model}}}$, are reduced via a pooling operation to single fixed-size vectors, $\mathbf{v}_{\text{graph}} \in \mathbb{R}^{d_{\text{model}}}$ and $\mathbf{v}_{\text{text}} \in \mathbb{R}^{d_{\text{model}}}$, respectively.

% These are concatenated to form the final feature vector, $\mathbf{v}_{\text{final}} = [\mathbf{v}_{\text{graph}} ; \mathbf{v}_{\text{text}}] \in \mathbb{R}^{2d_{\text{model}}}$, which is then passed to the regression head.

\textbf{Final prediction head}.
To account for the inherent uncertainty in PCE prediction, \method~is designed to predict a probability distribution that represents the possible range of PCEs.
In this work, we assume this distribution to be Gaussian.
Therefore, we add a head to \method~for mapping the fused representation to the parameters of the target probability distribution.
More precisely, the vector $\mathbf{v}_{\text{final}}$ serves as the input to the final prediction head, which is implemented as a Multi-Layer Perceptron (MLP).
%This head is responsible for mapping the fused representation to the parameters of the target probability distribution for our regression task.
This output layer produces two scalar values, which are interpreted as the predicted mean $\mu(x)$ and the standard deviation $\sigma(x)$ of the Gaussian distribution  $y \sim \mathcal{N}(\mu(x), \sigma(x)^2)$:
%and a raw value, $\sigma_{\text{raw}}$, that is subsequently transformed into the standard deviation, $\sigma$.
\begin{equation}
    \left[ \mu(x) ; \sigma(x) \right] = \operatorname{MLP}(\mathbf{v}_{\text{final}}).
\end{equation}

\subsection{Loss Function}

We frame the prediction task as a probabilistic regression to enable the model to predict the target PCE value and quantify the prediction uncertainty.
%the mean and standard deviation of the Gaussian distribution, we frame the prediction task as a probabilistic regression.
%to not only predict the target value but also to quantify its own prediction uncertainty, we frame the task as a probabilistic regression.
To this end, the model is trained to predict the parameters of a Gaussian distribution, $\mathcal{N}(\mu, \sigma^2)$, for each sample \cite{Bishop1994}.
%This approach, popularized in the context of Mixture Density Networks \cite{Bishop1994}, instead of producing a single point estimate, the model is trained to predict the parameters of a Gaussian (normal) distribution, $\mathcal{N}(\mu, \sigma^2)$, for each sample.
%The final regressor network outputs two values for each input: the predicted mean $\mu$ and a parameter that is transformed into the predicted standard deviation $\sigma$.

Formally, we train the model to minimize the Gaussian NLL of the true target values given the predicted distributions.
The probability density function for a single true target value $y$ given a predicted mean $\mu$ and variance $\sigma^2$ is given by:
\begin{equation}
    P(y | \mu, \sigma^2) = \frac{1}{\sqrt{2\pi\sigma^2}} \exp\left(-\frac{(y - \mu)^2}{2\sigma^2}\right).
\end{equation}
The training objective is to maximize the log-likelihood of the data, which is equivalent to minimizing its negative.
The NLL loss is therefore:
\begin{equation}
    \mathcal{L}_{\text{NLL}} =\frac{1}{B}\sum_{i=1}^{B} -\log P(y_i | \mu_i, \sigma_i^2) = \frac{1}{2B}\sum_{i=1}^{B}\left(\log(2\pi\sigma_i^2) + \frac{(y_i - \mu_i)^2}{\sigma_i^2}\right),
\end{equation}
where $B$ is the batch size. During optimization, the constant term $\frac{1}{2}\log(2\pi)$ can be disregarded as it does not affect the location of the minimum.
This yields the final loss function used in our work, averaged over a batch of samples:
\begin{equation}
    \mathcal{L} = \frac{1}{2B}\sum_{i=1}^{B}\left(\log(\sigma_i^2) + \frac{(y_i - \mu_i)^2}{\sigma_i^2}\right).
    \label{eq:nll_loss}
\end{equation}
This loss function consists of two terms. 
The term $\frac{(y - \mu)^2}{\sigma^2}$ is the variance-scaled squared error, which encourages the predicted mean $\mu$ to be close to the true value $y$.
The term $\log(\sigma^2)$ acts as a regularization term that penalizes the model for predicting excessively high variance.
{
\textcolor{black}
{Building on the NLL loss formulation, we could assess the calibration of predicted uncertainties by grouping samples into quantile bins and analyzing the relationship between predicted $\sigma$ and observed errors \cite{levi2022evaluating}. The details are provided in Appendix~\ref{app:calibration}.
}
\section{Experiments and Results}

In this section, we detail our experimental setup, covering datasets, implementation details, and evaluation metrics.
Then, we compare \method~against four baseline architectures designed for materials and device property prediction.
First, we consider the prior state of the art in perovskite device prediction, semantic GNN \cite{Aneesh2025}.
This model represents the entire solar cell as a heterogeneous graph and uses a GNN to learn inter-layer relationships from text-based material representations.
% \footnote{Currently, there is no public repository of Semantic GNN, so we implemented it with our best understanding.
% Our implementation was validated by achieving comparable performance on the original dataset.}.
Second, we adapt CrabNet \cite{Wang2021crabnet} for our device-level task, where individual layer representations are aggregated via mean pooling to form a single feature vector for the entire device, which is then used for the final PCE prediction.
We also implement a BERT+MLP baseline, where the textual descriptions of all device layers are encoded using a pre-trained MaterialsBERT model, and then these features are mean-pooled and fed to an MLP head for the regression task.
We also implement a CGCNN+BERT+MLP baseline.
This model first generates a global feature vector for the perovskite crystal using the CGCNN encoder and a separate feature vector for the device context by mean-pooling the BERT embeddings of the text layers.
These two vectors are then concatenated and passed to a final MLP head for prediction.
{
\color{black}
In addition, we introduce a LLM+Co-Attention baseline,  where chemical formulas describing each layer are processed by a large language model and fused with crystal features via a co-attention mechanism before prediction.
}
We also conduct a series of ablation studies on \textit{i}) the LLM, \textit{ii}) the geometric GNN,  \textit{iii}) the type of attention mechanism, and \textit{iv}) the type of loss function.
{
\color{black}
Finally, to further examine the impact of different data splitting strategies, we compare \method~against LLM+Co-Attention and semantic GNN under a group split based on device material configurations.
%the standard random split with a group split based on device material configurations for \textit{i}) \method, \textit{ii})  LLM+Co-Attention, \textit{iii}) semantic GNN.
}

%device architecture representation is combined through global mean pooling ,this vector is subsequently fed into a multi-layer perceptron (MLP) head for the final regression task.

%\subsection{Dataset}

\subsection{Experimental Setup}

\begin{table}[t]
    \centering
    \caption{Statistical summary of the dataset curation process, showing the reduction in total records and material diversity after filtering.}
    \label{tab:dataset_stats}
    \vspace{4pt}
    \begin{tabular}{lccc}
        \toprule
        \textbf{Metric} & \textbf{Original} & \textbf{Final} & \textbf{Reduction (\%)} \\
        \midrule
        Total records & 43,398 & 29,344 & 32.4 \\
        Unique perovskite formulas & 465 & 34 & 92.7 \\
        Unique ETL materials & 1,468 & 1,159 & 21.0 \\
        Unique HTL materials & 1,978 & 1,416 & 28.4 \\
        Unique back contacts & 290 & 244 & 15.9 \\
        Unique substrates & 194 & 159 & 18.0 \\
        \bottomrule
    \end{tabular}
\end{table}

\textbf{Dataset}.
%The data used in the study was constructed by integrating two open-source datasets: the Perovskite Database \cite{Jacobsson2022} and the Materials Project \cite{Jain2013}.
%The Perovskite Database served as a foundational source for device-level specifications.
% Our dataset combines the Perovskite Database \cite{Jacobsson2022} for device-level specifications with the Materials Project \cite{Jain2013}
% For each solar cell record, we extracted the textual descriptions of the functional layer, namely the HTL, ETL, perovskite absorber, back contact stack sequence, substrate stack sequence, and PCE.
We construct our dataset by combining device-level specifications from the Perovskite Database \cite{Jacobsson2022} with the Materials Project \cite{Jain2013}, extracting functional layer descriptions (HTL, ETL, absorber, back contact, substrate) and the corresponding PCE values.

%Our dataset combines the Perovskite Database \cite{Jacobsson2022} for device-level specifications with the Materials Project \cite{Jain2013}, extracting textual descriptions of functional layers (HTL, ETL, absorber, back contact, substrate) along with PCE values.
% \FloatBarrier
% \newpage 
\begin{wrapfigure}{r}{0.51\textwidth}
    \vspace{-10pt}
    \centering
    \captionof{table}{Training hyperparameters for the \method~model.}
    \label{tab:hyperparameters}
    \resizebox{0.51\textwidth}{!}{
    \begin{tabular}{lc}
    \toprule
    \textbf{Hyperparameter} & \textbf{Value} \\
    \midrule
    Hidden dimension (co-attention) & $64$ \\
    MLP layer dimensions & $[128, 64, 2]$ \\
    Co-attention layers & $3$ \\
    Normalization on co-attention $\gamma(\cdot)$ & $\operatorname{LayerNorm}$ \\
     Attention heads & $4$ \\
     Dropout rate & $0.2$ \\
     %Optimizer & AdamW \\
     Batch size & $16$ \\
     Learning rate (main modules) & $1 \times 10^{-4}$ \\
     Learning rate (MaterialsBERT) & $1 \times 10^{-6}$ \\
     %LR (CGCNN) & 0 (Frozen) \\
     Learning rate schedule & $\operatorname{Cosine}$ \\
     Warm-up epochs & $10$ \\
     Weight decay & $1 \times 10^{-5}$ \\
     Total epochs & $200$ \\
     Early stopping patience & $30$ \\
     \bottomrule
    \end{tabular}
    }
    %\vspace{-10pt}
\end{wrapfigure}

To incorporate atomic-level structural information, the Materials Project was cross-referenced to retrieve the corresponding crystallographic information.
This hybrid data collection strategy results in a multi-modal dataset where each sample is represented by two distinct data types: \textit{i}) a crystallographic information for the geometric construction of the crystal graph, and \textit{ii}) a set of textual strings describing the device context.
This filtering process, while essential for structural encoding, significantly refines the dataset as summarized in Table~\ref{tab:dataset_stats}. 
The total number of device records was reduced by 32.4\%, from an initial 43,398 to a final 29,344 data points after matching. 
More critically, this process substantially narrows the chemical space of the perovskite absorbers.
The requirement crystal structure in the Materials Project results in a 92.7\% reduction in the diversity of unique perovskite formulas, from 465 to 34.

\textbf{Metrics}.
%We evaluate the performance of \method~against several baseline models to demonstrate its effectiveness.
% We evaluate \method~using standard regression metrics including the R$^2$ score (coefficient of determination) and the mean absolute error (MAE).
% Additionally, we use the prediction interval coverage probability (PICP) \cite{Dewolf2023} to evaluate the calibration of the predicted uncertainty.
We evaluate \method~using standard regression metrics (R$^2$ and MAE), the Spearman's rank correlation coefficient (Spearman's $\rho$) to assess ranking capability, and the prediction interval coverage probability (PICP) \cite{Dewolf2023} to asses uncertainty calibration.
%with the prediction interval coverage probability (PICP) \cite{Dewolf2023}. 

%, the prediction interval coverage probability (PICP) \cite{Dewolf2023} was applied.

\textbf{Implementation details}.
%\textcolor{red}{...}
% We train \method~using the AdamW optimizer \cite{loshchilov2017decoupled}. 
% We employ a cosine learning rate scheduler with a warm-up period of $10$ epochs.
% The learning rates for different parts of the model were set individually, with the pre-trained CGCNN encoder being frozen during training.
% The dataset was split into training (80\%), validation (10\%), and test (10\%) sets.
% All experiments are repeated three times with different random seeds to ensure statistical robustness.
% All hyperparameters are detailed in Table \ref{tab:hyperparameters}.
We train \method~with the AdamW \cite{loshchilov2017decoupled} optimizer and a cosine learning rate schedule ($10$-epoch warm-up), freezing the pre-trained CGCNN encoder.
The dataset is split into 80\%/10\%/10\% for training, validation, and testing, and experiments are repeated with three random seeds.
Full hyperparameters are listed in Table \ref{tab:hyperparameters}.

\subsection{Comparison with the State of the Art}

\begin{table}[t]
    \centering
    \caption{Performance comparison of Solar-GECO with baseline models. Results are reported as mean $\pm$ standard deviation over three runs with different random seeds. The best results are highlighted in \textbf{bold}. Stars indicate that a baseline's performance is statistically significantly different from \method: $^{*}p < 0.05$ (significant difference from \method), $^{***}p < 0.001$ (highly significant difference from \method).}
    \vspace{4pt}
    \label{tab:performance_comparison}
    \begin{tabular}{lccc}
        \toprule
        \textbf{Model} & \textbf{R$^2$ Score} $\uparrow$ & \textbf{MAE} $\downarrow$ & \textbf{Spearman's $\rho$} $\uparrow$\\
        \midrule
        % Main Model
        %\midrule
        % Baseline Models
        BERT+MLP & $0.3863\pm 0.0043^{***}$ & $3.0436 \pm 0.0164^{***}$ & $0.5944\pm 0.0136^{*}$\\
        CGCNN+BERT+MLP & $0.4009 \pm 0.0067^{*}$ & $3.0111 \pm 0.0324 ^{*}$ & $0.6109 \pm 0.0047^{*}$\\
        CrabNet & $0.2090 \pm 0.0058^{***}$ & $3.3655 \pm 0.0072^{***}$ & $0.3807 \pm 0.0024^{***}$ \\
        LLM Co-attention & $ 0.4048 \pm 0.0049^{*}$ & $2.9812 \pm 0.0104^{*}$ & $0.6120 \pm 0.0003^{*}$ \\
        Semantic GNN & $ 0.3907 \pm 0.0105^{*}$ & $3.0668 \pm 0.0471^{*}$ & $0.5943 \pm 0.0071^{*}$ \\[0.2em]
        \cdashline{1-4}\\[-0.8em]
        \method & $\textbf{0.4179} \pm \textbf{0.0042}$ & $\textbf{2.9361} \pm \textbf{0.0179}$ & $\textbf{0.6192} \pm \textbf{0.0034}$\\
        \bottomrule
    \end{tabular}    
    % \vspace{0.5em}
    % \begin{flushleft}
    %     \footnotesize
    %     \textit{Significance levels from t-test vs. \method: $^{*}p < 0.05$,  $^{***}p < 0.001$}
    % \end{flushleft}
\end{table}

% \begin{wrapfigure}{r}{0.5\textwidth}
%     \vspace{-10pt}
%     \centering
%     \includegraphics[width=0.5\textwidth]{NeurIPS_AI4Mat_2025/figures/parity_plot_test_set.pdf}
%     \caption{Plot of predicted PCE versus true PCE for the original model on the test set. The diagonal line represents perfect prediction.}
%     \label{fig:pce_parity_plot}
%     \vspace{-15pt}
% \end{wrapfigure}

Table \ref{tab:performance_comparison} compares Solar-GECO with the baseline models.
Across both evaluation metrics, Solar-GECO consistently achieves the best performance, demonstrating significant improvements over existing approaches.
We verify the statistical significance of these results using t-tests against all baselines.
Stars in Table~\ref{tab:performance_comparison} mark cases where differences are significant ($p<0.05$ or $p<0.001$), confirming that the observed gains are robust rather than random fluctuations.

The largest performance gap in Table \ref{tab:performance_comparison} is relative to CrabNet, showing the limitations of composition-only methods for modeling complex, multi-scale device properties.
The BERT+MLP baseline benefits from contextual embeddings provided by the MaterialsBERT model, yet \method’s statistically significant gains show that explicitly modeling the interactions between atomic structures and device layers is more effective.
The performance metrics reveal a statistically significant advantage for the \method~over the CGCNN+BERT+MLP baseline, achieving a higher R$^2$ score (0.4179 vs. 0.4009) and a significantly lower MAE of 2.9361 compared to the baseline's 3.0111.

\begin{wrapfigure}{r}{0.6\textwidth}
    \vspace{-10pt}
    \centering
    \includegraphics[width=0.6\textwidth]{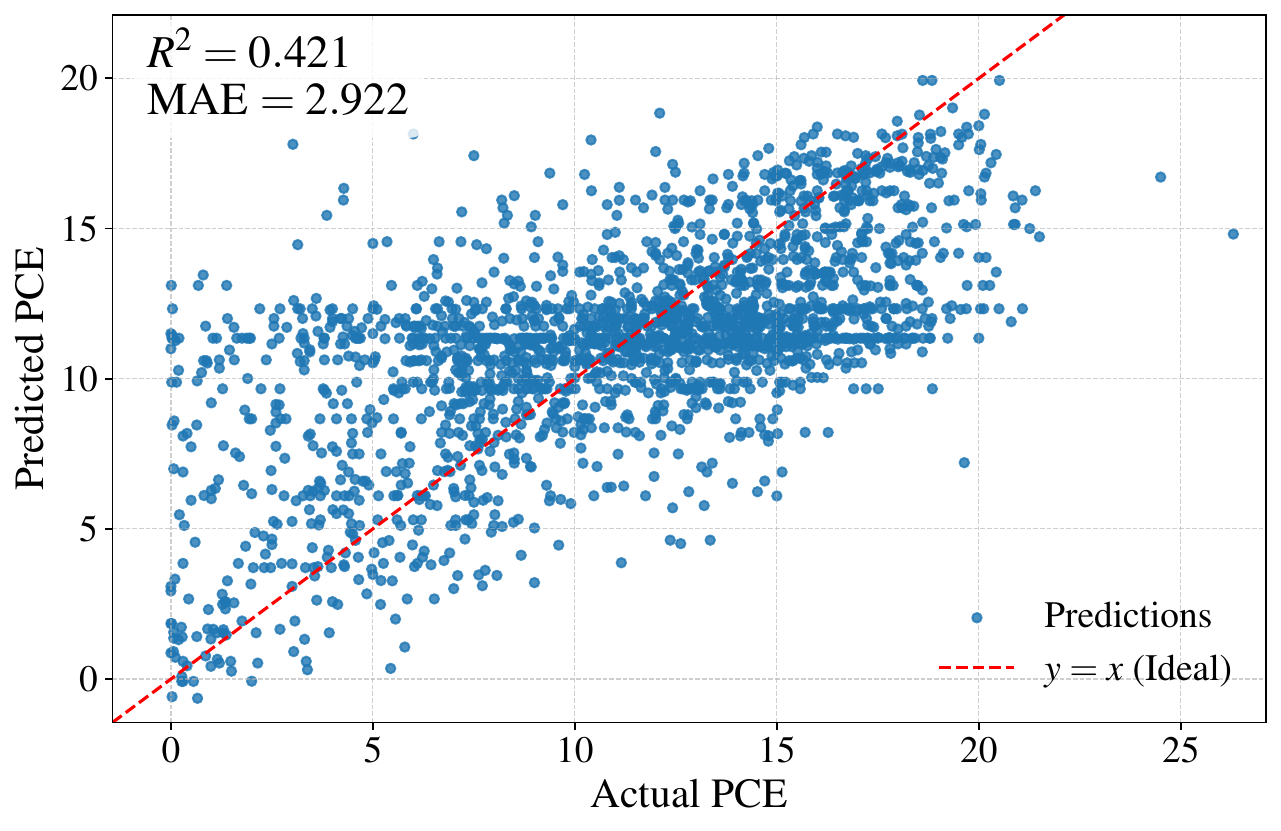}
    \caption{Plot of predicted PCE versus true PCE for the original model on the test set. The diagonal line represents perfect prediction.}
    \label{fig:pce_parity_plot}
    \vspace{-10pt}
\end{wrapfigure}
%\FloatBarrier
Furthermore, the improved performance of \method~over the BERT+MLP baseline highlights the significant benefit of incorporating the CGCNN to model the core perovskite layer.
This performance improvement can be also attributed to the integration of the co-attention module, since it helps to align the two modalities and capture interactions between device layers.
\method~also outperforms the semantic GNN, it reduces the MAE from 3.0668 to 2.9361 and improves R$^2$ from 0.3907 to 0.4179, highlighting the importance of incorporating geometric information from the crystal lattice, which text-based embeddings alone cannot capture.
%\method’s gains arise from its architecture: by encoding the perovskite’s 3D crystal structure with a geometric GNN, the model captures essential structure–property relationships that text-based methods miss.
In terms of ranking ability, our model \method~also demonstrates superior performance, achieving a Spearman's $\rho$ of 0.6192 $\pm$ 0.0034.
{
\color{black}
We also evaluate the LLM+Co-Attention baseline, which achieves an MAE of 2.9812 $\pm$ 0.0104, R$^2$ of 0.4048 $\pm$ 0.0049, and Spearman’s $\rho$ of 0.6120 $\pm$ 0.0003. While competitive, its performance is slightly below Solar-GECO, showing that co-attention improves modality alignment but lacks geometric crystal information for optimal accuracy.}
%The result of \method~is significantly better than all baselines, proving our method's excellent capability, which is crucial for candidate materials high-throughput screening.

% These improvements stem from the architectural design of Solar-GECO.
% The perovskite crystal is the core functional component of the solar cell, and its atomic geometry directly affects device efficiency.
% By using a geometric GNN, our model encodes the 3D structure of the absorber, capturing critical chemistry–geometry relationships that LLM-based methods overlook.

Figure~\ref{fig:pce_parity_plot} further illustrates our model performance (shown for one seed only) through a parity plot of predicted versus actual PCE values. 
Most predictions cluster tightly around the ideal diagonal, reflecting the model’s overall accuracy and stability across the test set.
However, deviations are more pronounced for low-PCE devices, where larger relative errors suggest that accurately predicting the PCE of underperforming architectures is inherently more difficult.
%This regime likely exhibits higher noise and variability in the experimental data, which may limit predictive resolution. 
Nevertheless, \method~maintains strong global consistency, reinforcing the robustness of our approach.

\begin{wrapfigure}{r}{0.56\textwidth}
    \vspace{-10pt}
    \centering
    \includegraphics[width=0.56\textwidth]{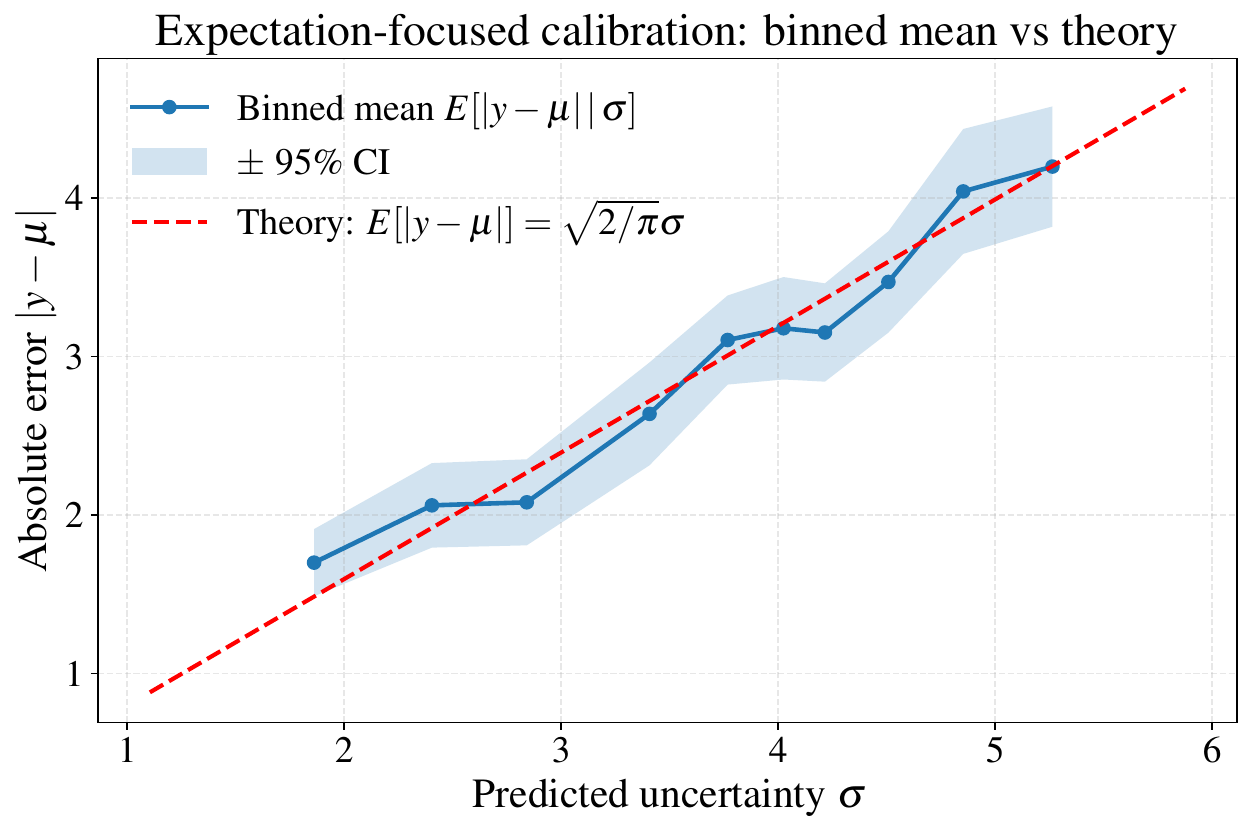}
    \caption{Calibration plot: quantile-binned mean $\overline{|e|}$ vs.\ $\overline{\sigma}$ with $95\%$ confidence intervals (CIs)  for the \emph{mean}, overlaid with the theoretical line $c\,\sigma$ , $c=\sqrt{2/\pi}$.}
    \label{fig:pce_sigma_plot}
    \vspace{-15pt}
\end{wrapfigure}
To evaluate the quality of the model's predicted uncertainty, we construct prediction intervals with a nominal confidence level of $95\%$ and calculate the PICP on the test set.
Our model achieved a PICP of $0.9593$.
This empirical coverage is in agreement with the $95\%$ nominal level (an absolute difference of $0.93\%$), which indicates that the predicted uncertainty is well-calibrated.
\textcolor{black}{
In Figure~\ref{fig:pce_sigma_plot}, we test the hypothesis that the expected absolute error $\mathbb{E}[|y-\mu(x)|]$ scales linearly with predicted uncertainty $\sigma(x)$ via $c=\sqrt{2/\pi}$. This is supported by the 95\% confidence interval (blue band), which reflects sampling variability. The theoretical line (red dashed) lying almost entirely within this band confirms that deviations from the empirical mean (blue line) are not statistically significant, indicating well-calibrated $\sigma$ values across the output range.
}

\subsection{Ablation Studies}

To assess the contribution of each component in \method, we perform several ablation experiments by systematically replacing key modules of our model.
The results of the ablation studies are shown in Tables \ref{tab:LLM_ablation}, \ref{tab:GNN_ablation}, \ref{tab:Attention_ablation}, and \ref{tab:Loss_ablation}.

First, we compare MaterialsBERT \cite{Gupta2022} with MatSciBERT \cite{gupta_matscibert_2022} as the text encoder in Table~\ref{tab:LLM_ablation}.
MaterialsBERT yields a slightly lower MAE (2.922 vs. 2.924), which might be attributed to its broader and more diverse pre-training corpus.
Second, we test the crystal graph encoder by replacing the pre-trained CGCNN \cite{Xie2018} with CHGNet \cite{deng_2023_chgnet} in Table~\ref{tab:GNN_ablation}. 
The performance drop suggests that representations optimized for bulk property prediction are less transferable to the downstream task of device-level PCE estimation, whereas CGCNN features seem to generalize more effectively.
Also, we replace the standard cross-attention with a gated contextual cross-attention mechanism \cite{ZHANG2024106553} in Table~\ref{tab:Attention_ablation}. 
This leads to a marked degradation in performance, indicating that the added parametric complexity was not beneficial for our dataset size and may have introduced overfitting.
Finally, we also perform an ablation on the loss function, comparing the standard mean squared error (MSE) with the probabilistic Gaussian NLL loss.
The results in Table~\ref{tab:Loss_ablation} show that both loss functions achieve an identical MAE of 2.922, while the Gaussian NLL loss yields a higher R$^2$ score (0.421 vs. 0.415).
Overall, the ablation results indicate that the combination of MaterialsBERT, CGCNN, and the vanilla attention mechanism provides the best trade-off between model complexity and predictive accuracy, supporting the overall design choices of \method.

\begin{table}
\parbox{.48\linewidth}{
\centering
    \caption{Ablation on the LLM encoder.}
    %\resizebox{0.5\columnwidth}{!}{
    \begin{tabular}{ccc}
       \toprule
       \textbf{Text Encoder} & \textbf{R² Score} $\uparrow$ & \textbf{MAE} $\downarrow$ \\
       \midrule
       MatSciBERT & \textbf{0.421} & 2.924 \\
       MaterialsBERT & \textbf{0.421} & \textbf{2.922} \\
     \bottomrule
    \end{tabular}
    %}
    \label{tab:LLM_ablation}
}
\hfill
\parbox{.5\linewidth}{
\centering
   \caption{Ablation on the geometric GNN encoder.}
   \begin{tabular}{ccc}
       \toprule
       \textbf{Geometric GNN} & \textbf{R² Score} $\uparrow$ & \textbf{MAE} $\downarrow$ \\
       \midrule
       CHGNet & 0.394 & 3.032 \\
       CGCNN & \textbf{0.421} & \textbf{2.922} \\
     \bottomrule
    \end{tabular}
    \label{tab:GNN_ablation}
}

\parbox{.48\linewidth}{
\centering
   \caption{Ablation on the attention mechanism.}
   \begin{tabular}{ccc}
       \toprule
       \textbf{Attention Function} & \textbf{R² Score} $\uparrow$ & \textbf{MAE} $\downarrow$ \\
       \midrule
       Gated & 0.372 & 3.108 \\
       Vanilla & \textbf{0.421} & \textbf{2.922} \\
       % Gated Cross-Attention & 0.372 & 3.108 \\
       % Vanilla Cross-Attention & \textbf{0.421} & \textbf{2.922} \\
     \bottomrule
    \end{tabular}
    \label{tab:Attention_ablation}
}
\hfill
\parbox{.5\linewidth}{
\centering
   \caption{Ablation on the loss function.}
   \begin{tabular}{ccc}
       \toprule
       \textbf{Loss Function} & \textbf{R² Score} $\uparrow$ & \textbf{MAE} $\leftrightarrow$ \\
       \midrule
       MSE & 0.415 & 2.922 \\
       Gaussian NLL & \textbf{0.421} & \text{2.922} \\
     \bottomrule
    \end{tabular}
    \label{tab:Loss_ablation}
}

\end{table}

{\color{black}
\subsection{Sensitivity Studies}
To analyze sensitivity, we adopt an alternative data splitting strategy based on device-layer material groups, in addition to the standard random split. Using this group split, we conduct experiments on \method, along with the Semantic GNN and LLM+Co-Attention baselines for comparison.
The results of this experiment are summarized in Table~\ref{tab:sensitivity_results}.}
\begin{wraptable}{r}{0.5\textwidth}
    \centering
    \vspace{-10pt}
    \caption{\color{black}
    Sensitivity analysis comparing \method~with Semantic GNN and LLM+Co-Attention under group split.}
    \label{tab:sensitivity_results}
    %\vspace{4pt}
    \begin{tabular}{lccc}
        \toprule
        \textbf{Model} & \textbf{R² Score} & \textbf{MAE} \\
        \midrule
        Semantic GNN & 0.3700 & 3.2740\\
        LLM+Co-Attention & 0.3374 & 3.2820\\
        \method & \textbf{0.3724} & \textbf{3.1271}\\
        \bottomrule
    \end{tabular}
    %\vspace{-20pt}
\end{wraptable}
{\color{black}
This analysis reveals that the group-based split, which is designed to test for the generalization to unseen material combinations, presents a significantly more challenging benchmark for all models.
Under this split, \method~demonstrates the highest robustness, achieving the best performance with the lowest MAE (3.1271) and the highest $R^2$ score ($0.3724$), which shows its current architecture also effective at generalizing to novel device-layer material groups compared to the baseline approaches.
}

\subsection{Limitations}

Despite its strong performance, \method~has some limitations that suggest directions for future work.
First, the geometric data for the perovskite absorber is primarily obtained from the Materials Project \cite{Jain2013}, which constrains the diversity of available crystal structures and may limit exposure to less-represented materials.
Moreover, the current framework focuses on compositional and structural features but does not explicitly account for fabrication-related parameters, such as annealing temperature or deposition methods.
These parameters are known to strongly influence device performance, so considering them in the prediction of PCE is an important step towards having a reliable machine learning methodology that could be later used for exploration purposes.
% Additionally, we attempted to include fabrication process information for the perovskite layer on the text side, expanding the text input from $\mathbb{R}^{4 \times d_{\text{bert}}}$ to $\mathbb{R}^{5 \times d_{\text{bert}}}$. However, this resulted in only marginal performance improvement, indicating that more effective strategies for integrating process-related features remain an open research question.
{
\color{black}Another challenge we observed is the low PCE bias , balancing strategies such as adaptive sampling, importance weighting, or contrastive loss could alleviate this issue.
}

% Despite its strong performance, our study has several limitations that offer avenues for future research:
% \textbf{Data Limitations:}The geometric data for the perovskite layer is primarily sourced from Material Project databases, this dependency limits the overall diversity of our structural data.

% \textbf{Model Limitations:}The current framework focuses on compositional and structural aspects but does not explicitly incorporate fabrication process parameters (e.g., annealing temperature, deposition methods). These parameters are known to have a profound impact on final device efficiency.Also the more advanced attention mechanism could be introduced to improve model's generalization ability.

\section{Conclusions}

In this work, we introduced \method, a multi-modal geometric deep learning framework for predicting the PCE of perovskite solar cells. 
By combining a geometric GNN encoding of the perovskite crystal structure with LLM-based embeddings of device layers, and fusing them through a co-attention module, our model captures both intra- material properties and inter-device interactions. 
This representation leads to state-of-the-art accuracy, outperforming established baselines such as CrabNet and semantic GNN.
Our results demonstrate that explicitly incorporating crystal geometry alongside device context is essential for modeling complex, multi-scale perovskite devices. 
Beyond accuracy, \method~provides a practical tool for screening candidate device architectures, accelerating the discovery of efficient perovskite solar cells.

Future work will proceed on two fronts. First, we will integrate fabrication parameters, such as deposition procedures available in the Perovskite Database, as explicit input features.
Second, we will seek to expand the structural data diversity.
These combined efforts hold promise for further improving the model's generalization and applicability in the discovery of new perovskite devices.

\bibliography{main}
\bibliographystyle{ieeetr}

% \section*{References}

% References follow the acknowledgments in the camera-ready paper. Use unnumbered first-level heading for
% the references. Any choice of citation style is acceptable as long as you are
% consistent. It is permissible to reduce the font size to \verb+small+ (9 point)
% when listing the references.
% Note that the Reference section does not count towards the page limit.
% \medskip

% {
% \small

% [1] Alexander, J.A.\ \& Mozer, M.C.\ (1995) Template-based algorithms for
% connectionist rule extraction. In G.\ Tesauro, D.S.\ Touretzky and T.K.\ Leen
% (eds.), {\it Advances in Neural Information Processing Systems 7},
% pp.\ 609--616. Cambridge, MA: MIT Press.

% [2] Bower, J.M.\ \& Beeman, D.\ (1995) {\it The Book of GENESIS: Exploring
%   Realistic Neural Models with the GEneral NEural SImulation System.}  New York:
% TELOS/Springer--Verlag.

% [3] Hasselmo, M.E., Schnell, E.\ \& Barkai, E.\ (1995) Dynamics of learning and
% recall at excitatory recurrent synapses and cholinergic modulation in rat
% hippocampal region CA3. {\it Journal of Neuroscience} {\bf 15}(7):5249-5262.
% }

%%%%%%%%%%%%%%%%%%%%%%%%%%%%%%%%%%%%%%%%%%%%%%%%%%%%%%%%%%%%

\newpage
\appendix

{\color{black}
\section{Uncertainty Calibration Analysis Details}
\label{app:calibration}
To assess the calibration of predicted uncertainties, we group samples into quantile bins and analyze the relationship between predicted standard deviation $\sigma(x)$ and observed errors, which follows for each input $x$, a predictive Gaussian distribution 
$Y\mid X=\mathbf{x} \sim \mathcal{N}\!\bigl(\mu(\mathbf{x}),\,\sigma^2(\mathbf{x})\bigr)$,. Given a dataset $\mathcal{D}=\{(\mathbf{x_i},y_i)\}_{i=1}^M$,we construct N quantile bins on the scalar $\{\sigma_i \}_{i=1}^M$,let $\{\mathcal{I}_b\}_{b=1}^N$ be a partition of $\{1,\dots,M\}$ such that if $i\in\mathcal{I}_b$ and $j\in\mathcal{I}_{b'}$ with $b<b'$, then $\sigma_i\le \sigma_j$ , denote $n_b=|\mathcal{I}_b|$.
 For each bin $b$, we compute the average predicted uncertainty and the average absolute error:
\begin{equation}
\overline{\sigma}_b := \frac{1}{n_b}\sum_{i\in\mathcal{I}_b}\sigma_i,\qquad
\overline{|e|}_b := \frac{1}{n_b}\sum_{i\in\mathcal{I}_b}|e_i|,
\end{equation}

Define the residual $e := y-\mu(\mathbf{x})$ and the conditional residual $\varepsilon := y-\mu(\mathbf{x})\mid \mathbf{x}$. Assume the conditional residual is Gaussian:
\begin{equation}
\varepsilon \;\big|\; \mathbf{x} \;\sim\; \mathcal{N}\bigl(0,\,\sigma(\mathbf{x})^2\bigr).
\end{equation}

Equivalently, write $\varepsilon = \sigma(\mathbf{x})Z$ with $Z\sim\mathcal{N}(0,1)$. Then:
\begin{equation}
\label{eq:Eabs}
\mathbb{E}\!\left[\,|y-\mu(\mathbf{x})| \,\middle|\, \mathbf{x} \right]
= \mathbb{E}\!\left[\,|\varepsilon| \,\middle|\, \mathbf{x} \right]
= \sigma(\mathbf{x})\,\mathbb{E}|Z|
= c\,\sigma(\mathbf{x}),
\end{equation}
where the half-normal mean constant is:
\begin{equation}
\mathbb{E}|Z|
=\int_{-\infty}^{+\infty} |z| \,\frac{1}{\sqrt{2\pi}}e^{-z^2/2}\,dz
= 2\int_{0}^{+\infty} z\,\frac{1}{\sqrt{2\pi}}e^{-z^2/2}\,dz
= \sqrt{\frac{2}{\pi}}.
\end{equation}

The expected absolute error conditioned on $\mathbf{x}$ scales linearly with the predicted uncertainty $\sigma(\mathbf{x})$ with slope $c=\sqrt{2/\pi}$. We plot points $\bigl(\overline{\sigma}_b,\,\overline{|e|}_b\bigr)$ and overlay the theoretical line $c\,\sigma(\mathbf{x})$ from \eqref{eq:Eabs}. 

To express the uncertainty of the bin mean, we draw a $95\%$ confidence interval for the mean using the standard error:
\begin{equation}
\mathrm{SE}_b \;=\; \frac{s_b}{\sqrt{n_b}}, 
\quad \text{where}\quad
s_b^2 = \frac{1}{n_b-1}\sum_{i\in\mathcal{I}_b} \Bigl(|e_i|-\overline{|e|}_b\Bigr)^2,
\end{equation}
and the band $\overline{|e|}_b \pm 1.96\,\mathrm{SE}_b$.
}

% \section{Technical Appendices and Supplementary Material}
% Technical appendices with additional results, figures, graphs and proofs may be submitted with the paper submission before the full submission deadline (see above), or as a separate PDF in the ZIP file below before the supplementary material deadline. There is no page limit for the technical appendices.

%%%%%%%%%%%%%%%%%%%%%%%%%%%%%%%%%%%%%%%%%%%%%%%%%%%%%%%%%%%%

\end{document}